# Mesh Conflation of Oblique Photogrammetric Models using Virtual Cameras and Truncated Signed Distance Field

Shuang Song, *Student Member, IEEE* and Rongjun Qin, *Senior Member, IEEE*

*Abstract*— **Conflating/stitching 2.5D raster digital surface models (DSM) into a large one has been a running practice in geoscience applications, however, conflating full-3D mesh models, such as those from oblique photogrammetry, is extremely challenging. In this letter, we propose a novel approach to address this challenge by conflating multiple full-3D oblique photogrammetric models into a single, and seamless mesh for high-resolution site modeling. Given two or more individually collected and created photogrammetric meshes, we first propose to create a virtual camera field (with a panoramic field of view) to incubate virtual spaces represented by Truncated Signed Distance Field (TSDF), an implicit volumetric field friendly for linear 3D fusion; then we adaptively leverage the truncated bound of meshes in TSDF to conflate them into a single and accurate full 3D site model. With drone-based 3D meshes, we show that our approach significantly improves upon traditional methods for model conflations, to drive new potentials to create excessively large and accurate full 3D mesh models in support of geoscience and environmental applications.**

*Index Terms—model conflation, oblique photogrammetry, truncated signed distance field (TSDF), LiDAR, virtual cameras.*

## I. INTRODUCTION

Typical geoscience applications rely on large-scale and wide-area 2.5D digital surface models (DSM) to characterize the earth surfaces. As a common practice, a large DSM is produced through stitching or conflating multiple individual and smaller DSMs, which could be generated from means such as photogrammetry, LiDAR, and SAR (Synthetic Aperture Radar) interferometry. In recent years, *Oblique Photogrammetry* is attracting increasing attention, due to its capability to characterize sites in full 3D, which greatly facilitates applications such as digital twin, disaster responses to landslides, and earthquakes that large-scale site models with detailed façade geometry. However, unlike model conflation for 2.5D raster DSM, conflating full 3D site models is extremely challenging, as these models are often represented in full-3D triangle meshes, whose topology among vertices is intricate, and oftentimes associated with complex manifold geometry. This is made further even more challenging when these individual mesh models are from different sources, with different resolutions, accuracy, and level of uncertainty. Therefore, when presenting multiple mesh models, the current practices simply overlay these models for visualization, and at most, moderate the mesh conflation process through simple averaging [1]–[3], followed by manual or semi-automatic mesh editing through Boolean operations, thus leaving mesh conflation a largely underserved task. Recent advances in 3D scene representations, such as through implicit surfaces, occupancy grid, signed distance field (SDF), and truncated signed distance field (TSDF) [4], have made it possible to manipulate 3D assets under continuous fields. Among these, TSDF has been a well-supported 3D implicit representation model in the Computer Graphics (CG), due to its simplicity in representing compact spaces, its scalability for large scene processing due to the locality of the updating process and marching cubes [5], and its capability to set forth 3D geometry fusion/conflation frameworks.

Therefore, in this letter, we present a novel approach that uses a virtual camera [6] induced TSDF to fuse multiple mesh models into a single and coherent mesh. Specifically, our approach fills a few gaps of prior works in view-based large-scale mesh fusion/conflation: 1) we propose a novel panoramic virtual camera field to minimize the chances of occlusion at complex objects. 2) we introduce a scheme concerning non-uniform weights with respect to the quality of individual meshes, thus it provides a more reliable and accurate means for mesh model conflation. 3) we adapt the ground-plane assumption oblique photogrammetric models to automate the virtual camera placement.

## II. RELATED WORKS

*Mesh conflation* aims to combine mesh models from different sources or scans into a single and coherent mesh model [3]. As mentioned earlier, this is a rather underserved task, yet is still highly relevant to a number of existing approaches [7]–[12]. Remeshing is a family of methods in computer graphics, which has been used to create a new mesh

This manuscript is initially submitted on 2023-03-12.

This work was supported in part by the Office of Naval Research, Award No. N00014-20-1-2141 & N00014-23-1-2670

*Corresponding author: Rongjun Qin*

Shuang Song is with the Department of Civil, Environmental and Geodetic Engineering, The Ohio State University, Columbus, USA (e-mail: song.1634@osu.edu).

Rongjun Qin is with the Department of Civil, Environmental and Geodetic Engineering and the Department of Electrical and Computer Engineering, and Translational Data Analytics Institute, The Ohio State University, Columbus, USA (e-mail: qin.324@osu.edu)

Color versions of one or more of the figures in this article are available online at http://ieeexplore.ieee.org

with a desired topology conditioned on the original surface geometry. It is categorized into two types: explicit remeshing and implicit remeshing. Explicit remeshing primarily preserves the 3D vertices and generates triangular networks subjected to desired constraints, such as number of faces and conformation [13]. Algorithms in this category are primarily built based on heuristics and can be hardly extended to meshes representing complex objects [14]; The implicit remeshing first converts the original mesh to an intermediate volumetric representation, such as UDF (unsigned distance field), SDF (signed distance field), TSDF (truncated SDF), and then extracting a new mesh from the volumetric representation [5]. Most of these volumetric representations favor close surfaces [15], while being unsuitable for photogrammetric mesh models, which are typically open, with unknown manifoldness, and oftentimes with holes and artifacts. TSDF was considered as a useful tool to address such challenges, for example, TSDF can be built based on individual views to minimize the topological complexity in fusion and has been widely used in surface reconstruction applications [16]. Yet this approach requires known image views to guide the fusion and conflation process, which might not be always available.

## III. PROPOSED METHOD

We present a novel approach that directly conflates full-3D mesh models based on TSDF (Section III-A), through means of virtual cameras. Our approach does not require known image views and can benefit from the efficiency of TSDF presentations.

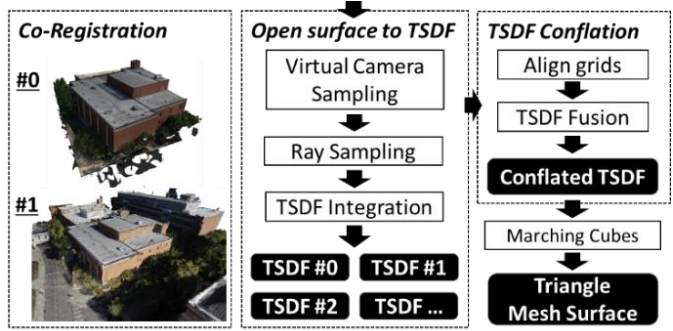

**Fig. 1. The workflow of our proposed mesh conflation approach.**

Fig. 1 presents the workflow of our proposed method. In our approach, we assume well-registered meshes [17] and start with sampling virtual cameras (Section III-B) in the outer space defined by mesh surfaces. Then we generate rays for each pixel originating from the virtual cameras and find the first intersection with the mesh surface. Since the mesh is a continuous surface, this process can yield pixel-wise depth values as the depth images. Then we integrate these rays into view-dependent TSDF and conflate them in the object space leveraging weights of different meshes based on their respective local geometry. This process is done for each view to update the TSDF field in the object space. Finally, we extract triangle meshes from the implicit surface representation using marching cubes methods [5].

### A. Open surface to TSDF

Signed Distance Function (SDF) defines a function $D(x)$ of the 3D space, where its function value at a 3D location $x$ defines the Euclidean distance of this $x$ location to the actual surface, and the sign indicates if $x$ is inner or outer the surface [18]. In a fusion framework, it usually couples with a weight function $W(x)$ defining the importance of a point at fusion. TSDF is a more memory-efficient version of SDF, since it truncates the field only near the surfaces at a bandwidth $m$ (see equation (1-3)). This feature makes it suitable for large-scale applications with limited resources. The TSDF can be easily constructed from closed surfaces using sweep [15] and flood-filling [19] algorithms. Yet, these methods do not apply to open surfaces, as flood-filling yields nothing due to the presence of a single connected components.

The method of [4] performs TSDF integration given a few depth images. Basically, the method presupposes iterative updating of TSDF distance $D(x): \mathbb{R}^3 \rightarrow \mathbb{R}$ and weight $W(x): \mathbb{R}^3 \rightarrow \mathbb{R}$ functions. Both are real-valued functions defined for every voxel of the voxel grid. They are initialized with zeros and updated for each ray (Equation (1-3)) in every depth image.

$$r_i(t) = o_i + t \cdot v_i . \tag{1}$$

$$d(r_i(t)) = \max(-m, \min(m, t_i - t)), \tag{2}$$

$$w(r_i(t)) = \begin{cases} C, & \text{if } |t_i - t| \le m \\ 0, & \text{otherwise} \end{cases}, \tag{3}$$

where $o_i \in \mathbb{R}^3$ is the position of camera center, $v_i \in \mathbb{R}^3$ is the normalized direction of the ray, $t$ is the length of the ray, $d(x)$ and $w(x)$ define TSDF along the ray $r_i$, $t_i$ is the distance from ray to nearest surface, and a distance $m$ defines the vicinity of the surface. In common practice, $m$ is set to 3 times the voxel width. Then, the TSDF can be update using (3) and (4) in [4].

In our case, the weight function [4] is not suitable for a wide band width $m$. The problem is resolved by introducing adaptive weights (Section III-C).

### B. Virtual ray sampling

Converting explicit mesh models to TSDF requires ray sampling through the known geometry to the view space. However, typically, view space information does not exist (e.g., LiDAR data), or is not stored in mesh models. We hence propose to construct a virtual view field, and subsequently, design casted rays between the surface and these views, including their origin, direction, and length.

*Virtual camera placement for determining ray origins.* This virtual camera placement problem [6], [20] considers sampling a small number of observing cameras, such that their view space completely cover the scene with optimal angles. To maximize the coverage, we propose to place virtual cameras with a panoramic field of view (FoV), and since it covers all directions of concern. We propose a heuristic virtual camera placement algorithm: it first builds a coarse occupancy grid from the mesh model (green wireframe in Fig. 2). Then, the algorithm (shown in **Algorithm 1Error! Reference source not found.**) samples virtual camera centers based on the occupancy grid. Since outdoor scenes mostly follow a ground-plane assumption (XOY plane sets the ground, and Z relays the height), the camera placement starts with the highest cell for each XOY cell (top height layer of the voxel grid $G_z$).

To avoid empty cells (holes), instead of placing the camera for each cell of $G_z$, we apply a window size $\phi$ (set as 3 voxels) and look for the highest cell within the window. For cells at facades (large height difference), we sample along the vertical direction per voxel (For details please see **Algorithm 1**).

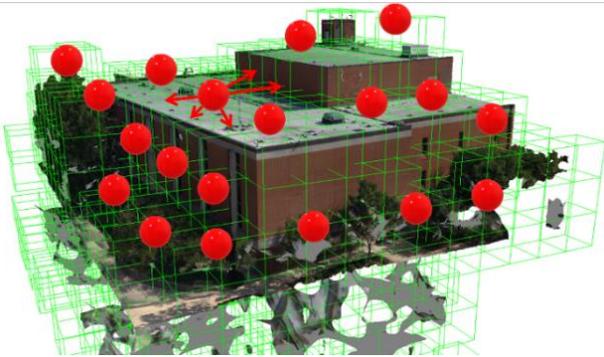

**Fig. 2. Virtual panoramic cameras (red spheres) based on the occupancy of the aerial model.**

---

**Algorithm 1** Virtual Camera Placement.
**input** $G_z$: Top height layer of Voxel Grid.
**input** $\phi$: window size
1: **procedure 1:** find_high_cell($G_z$, i, j, $\phi$)
2: z = -1
3: **for** ii=i-$\phi$; ii<i+$\phi$; ii++ **do**
4:     **for** jj=j-$\phi$; jj=j+$\phi$; jj++ **do**
5:         z = max(z, $G_z$[ii,jj])
6: **return** z
7:
8: **procedure 2:** sample_camera_centers ($G_z$, $\phi$)
9: // $G_z$: p,q matrix. p,q for x,y index
10: **for** i=0; i<p; i++ **do**
11:     **for** j=0; j<q; j++ **do**
12:         begin_k[i,j] = find_high_cell($G_z$, i, j, $\phi$)
13: **for** i=0; i<p; i++ **do**
14:     **for** j=0; j<q; j++ **do**
15:         end_k[i,j] = max(begin_k[i-$\phi$:i+$\phi$,j-$\phi$:j+$\phi$])+1
16:         **for** k=begin_k[i,j]; k<end_k[i,j]; k++ **do**
17:             **yield** i,j,k  // i,j,k refers to camera location

---

With the determined virtual cameras, the *Ray direction* can be sampled from the unit sphere $\mathcal{S}^2$ (i.e., panoramic FoV) of each virtual camera. This is typically done by sampling rays using regular latitude and longitude grids, while such a method can lead to a higher concentration of samples near the poles. Instead, we use the Fibonacci lattice method [21], which was proven to be more uniformly distributed on the $\mathcal{S}^2$.

*Ray length*, defined as t in (1), describes the traveling distance from the origin to the object's surface. Given the sampled virtual camera position $o_i$ and direction $i$, $t$ is determined by the intersection test between the ray and the geometry. This is supported by modern graphics-card based raytracing, in which billions of rays can be tested in a second.

### C. TSDF based Conflation

To adapt the model conflation application into the TSDF integration framework [4], we proposed additional modifications to accommodate data with different quality, by using adaptive weights associated with the bandwidth of the TSDF (Equation (2-3)): by definition, when converting the mesh model into TSDF, only the signed distance (Equation (2)) within the truncated band has non-zero values, so as the weights (Equation (3)). Zero weights outside the truncated band may cause ambiguities. Therefore, the bandwidths (as part of the weight definition) of TSDF from different sources of mesh can be adjusted in referencing to the quality of different meshes, which can be determined based on prior knowledge about the dataset itself (e.g., resolution and accuracy). A weighted average over these TSDFs can be applied to achieve the final conflation.

In addition, since the TSDF is discontinuous at truncated boundary, in certain cases, it generates a branched line as illustrated in Fig. 3(d) on occasions when non-convex shapes are formed when two triangles intersect. When in 3D it will yield artifacts causing missing details.

We proposed a novel approach to further improve the adaptiveness of our method on the truncated band: First, we define the maximal bandwidth based on the widest uncertainty bounds of the input datasets. Then, instead of using this fixed bandwidth value, we gradually reduce the bandwidth from one side (in our case we take the negative side) to ensure the bandwidth of one triangle does not interfere with another. This process is shown in Fig. 4: based on the surface point $P$, we extend its ray to $\lambda$, to which the negative bandwidth is defined. The final conflated surface can then be extracted with well-known marching cubes algorithms [5]. The resulting conflation output is presented in Section IV-B.

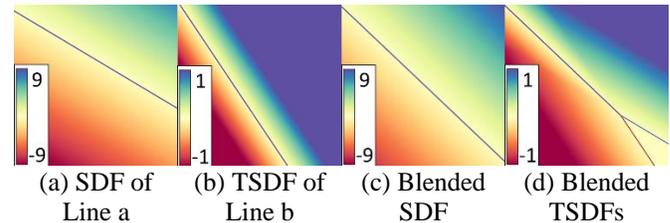

(a) SDF of Line a    (b) TSDF of Line b    (c) Blended SDF    (d) Blended TSDFs

**Fig. 3. Illustration of truncated boundary in terms of TSDF fusion. Color legends represent normalized units.**

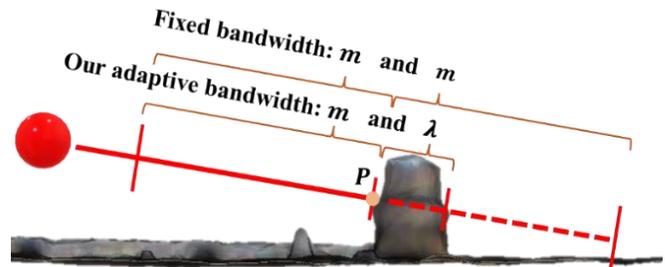

**Fig. 4. Adaptive bandwidth determination based on surface bounds. The areas crossed by the solid line represent the positive distance (outside of the object) while the dashed line indicates the negative distance (inside of the object). A fixed bandwidth in the classic TSDF would result in the contamination of empty regions, whereas our adaptive bandwidth does not suffer from this issue.**

## IV. EXPERIMENTS

### A. Implementation

Our implementation makes use of the OpenVDB [22], a B+trees based data structure and library for efficient storing and processing large-scale volumetric data. VDBFusion [23] is a utility library that implements the TSDF integration pipeline (Section III-A) using OpenVDB as the data structure. We verified the effectiveness of our approach by testing it on two oblique photogrammetric meshes that were captured by a drone at different altitudes representing different resolution and quality. The reconstructed meshes are of varying levels of uncertainty, resolution, and coverage. Our experiments were conducted on a Windows 10 workstation equipped with an Intel i7-8700, 32GB RAM, and the maximal bandwidth was set as 5 meters.

### B. Experiments on mesh model conflation

In this experiment, we demonstrate the conflation of two mesh models with different resolution and accuracy. The first model encompasses a larger area, yet it exhibits lower resolution. With 81K vertices and 167K triangles, it spans an area of 24,422 $m^2$. Conversely, the second model comprises 514K vertices and 1M triangles, covering an area of 2,961 $m^2$, which is approximately 12% of the first model. The second model is characterized by finely detailed structures, as illustrated in Fig. 5a.

We compare the performance of three methods: simple overlay, a remeshing algorithm called APSS (Algebraic Point Set Surfaces) [24], and our approach using both qualitative and quantitative metrics. APSS is a local method based on Moving Least Squares (MLS), which computes an implicit function from local point structures and converts these to a triangle mesh using the marching cubes algorithm [5]. The reference data is generated by photogrammetry software through a single and higher resolution collection of the region. It has 21M vertices and 9M triangles.

As shown in Fig. 5(a), a simple overlay method results in meshes intersecting with each other due to the data uncertainties (yellow and grey represent two mesh models). Since the simple overlay keeps individual meshes in their original form, the resulting topology of the meshes is incoherent. We also compared the APSS implemented in MeshLab and decimate it to have the same number of faces as our method, as shown in Fig. 5(b). While this method has merged the surfaces and preserved high-resolution mesh, it generates holes indicating the bad topology quality in overlapping regions. In contrast, Fig. 5(c) shows our conflated mesh, which is coherent, smooth, complete, and compact.

We analyze the distance between results and the reference mesh in two ways: the *mean distance* to the reference mesh, as well as the *F-Score* [25]. F-score is the harmonic mean between precision and recall, reflecting the proximity of the reconstructed points to the reference mesh and its completeness. Using a threshold of 0.5 meters (equal to the voxel size in our experiments and the highest resolution of source meshes), differences smaller than 0.5m are considered acceptable. Table I verifies that our method achieves the best result, yielding comparable accuracy to the original mesh, with better F-score.

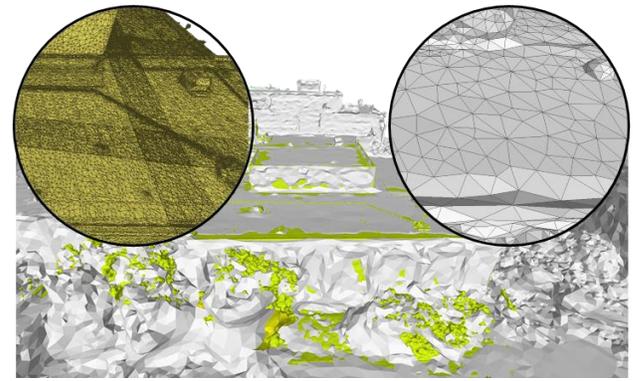

(a) Simple overlay method.
High-res dataset (yellow) and low-res dataset (grey).

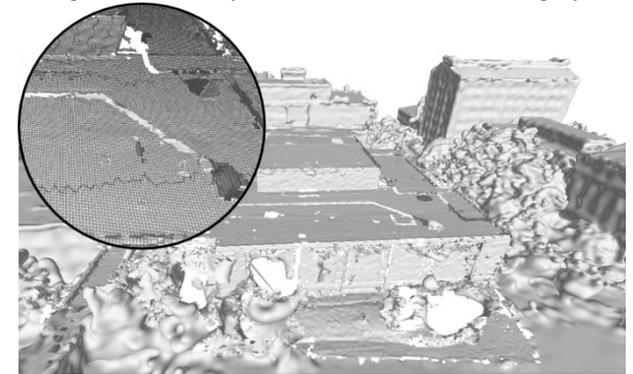

(b) APSS remeshing method in MeshLab [24].

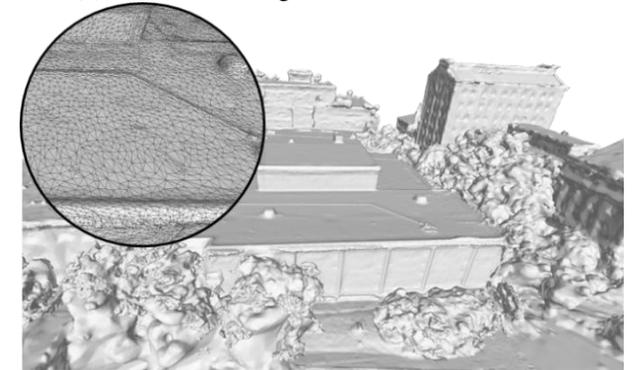

(c) Our conflated mesh.

**Fig. 5. Qualitative comparison of our conflation method with APSS [24] remeshing method.**

TABLE I
QUANTITATIVE RESULTS OF MODEL CONFLATION METHODS
COMPARING WITH THE REFERENCE MESH

| Methods | Mean distance | F-score | # Vertices | # Faces |
|---------|--------------|---------|-----------|---------|
| Overlay | **0.488 m** | 0.364 | 594 K | 1.2 M |
| APSS | 0.514 m | 0.365 | 569K | 1.1M |
| Ours | 0.503 m | **0.367** | 773K | 1.1M |

### C. Computational Complexity

The computational complexity of our method is relevant to several steps: ray casting, TSDF conflation, and marching cubes. Ray casting utilizes highly optimized methods, with a complexity that could be $O(N * log(M))$ or even lower, where $N$ is the number of rays and $M$ represents the number of triangles [26]. Leveraging OpenVDB, which can operate voxels in $O(1)$, the complexity of TSDF fusion process used by our proposed method is $O(N)$, as each ray could potentially influence a constant number of grid cells. The marching cubes algorithm, which scans over the activated voxels (truncated band) and visits only their local neighbors, carries a complexity of $O(P)$, where $P$ is the total number of voxels in the truncated band. This count correlates with the surface area of the input meshes. To sum up, our proposed method demonstrates better than linear complexity with respect to $N$, $M$, and $P$, offering significant scalability for larger datasets.

## V. CONCLUSION

In this letter, we presented a novel approach for conflating full-3D mesh models in support of large site modeling, an underserved task in oblique photogrammetry. Our approach adapts a novel virtual panoramic camera concept under a TSDF implicit surface representation. Our preliminary results demonstrated that the conflated model from our method excels existing methods both visually and statistically for complex scenes. It is 4 times more compact and 1.9% more accurate than a typical remeshing method with a coherent topology. As compared to existing practice that presents the earth surface in 2.5D raster, the approach developed in this letter has the potential to enable future full-3D mesh modeling for large area by leveraging data from different sources including photogrammetry and LiDAR data. However, it is important to note that our method currently supports only the exterior surface of geospatial models due to the assumption made in virtual camera sampling.